\documentclass[conference]{IEEEtran}

\IEEEoverridecommandlockouts
\usepackage{cite}
\usepackage{amsmath,amssymb,amsfonts}
\usepackage{algorithmic}
\usepackage{graphicx}
\usepackage{textcomp}
\usepackage{xcolor}
\usepackage{booktabs}
\usepackage{tabularx}
\usepackage{tikz}
\usetikzlibrary{arrows.meta,positioning,fit,backgrounds,calc,shapes.geometric}
\usepackage{url}
\graphicspath{{figs/}{images/}{./}}

\def\BibTeX{{\rm B\kern-.05em{\sc i\kern-.025em b}\kern-.08em
		T\kern-.1667em\lower.7ex\hbox{E}\kern-.125emX}}

\begin{document}
	
	\title{A Multi-Task Deep Learning Framework for\\ Real-Time
		Intelligent Video Surveillance with\\
		Temporal Event Validation}
	
	\author{
		\IEEEauthorblockN{1\textsuperscript{st} Estera Dumitru}
		\IEEEauthorblockA{\textit{Faculty of Information Systems and Cyber Security} \\
			\textit{Military Technical Academy ``Ferdinand I''}\\
			Bucharest, Romania \\
			estera.dumitru@mta.ro}
		\and
		\IEEEauthorblockN{2\textsuperscript{nd} Stelian Spînu}
		\IEEEauthorblockA{\textit{Faculty of Information Systems and Cyber Security} \\
			\textit{Military Technical Academy ``Ferdinand I''}\\
			Bucharest, Romania \\
			stelian.spinu@mta.ro
			}
		
	}
	
	\maketitle
	
	\begin{abstract}
		Modern video surveillance systems generate far more video streams than human operators can effectively monitor, making automated analysis essential for timely detection of security events. This paper presents a unified multi-task deep-learning framework that simultaneously performs face recognition with zone-based authorization, automatic license plate recognition, weapon detection, fire and smoke detection, and human action recognition on a shared GPU platform. Among the integrated modules, two task-specific deep-learning models are proposed in this work to address scenarios that are insufficiently represented in publicly available datasets: a single-class weapon detector fine-tuned on a merged and relabeled dataset, achieving a mean average precision (mAP@0.5) of 0.947, and a SlowFast-R50 action recognition model trained on a purpose-built vandalism dataset comprising 614 video clips, achieving 94.33\% classification accuracy. To improve robustness in continuous video, all detection modules are integrated into a temporal event-validation architecture based on multi-frame confirmation, confidence-weighted voting, and cascaded filtering, transforming frame-level predictions into reliable security events. Each module is evaluated independently on established public datasets (LFW, D-Fire, FIRESENSE, and UCF-Crime), followed by integrated end-to-end system evaluation. The proposed temporal validation strategy reduces the fire and smoke false-alarm rate from 52\% to 4\% and improves video license plate exact-match accuracy from 66.7\% to 81.8\%, while the complete framework maintains real-time operation with a per-frame latency below 100 ms on commodity hardware. These results demonstrate that combining specialized deep-learning models with temporal event validation provides an effective and practical solution for reliable real-time intelligent video surveillance.
	\end{abstract}
	
	
	\section{Introduction}
	
	Video surveillance has moved from a specialized military and industrial tool to a
	routine part of urban and residential infrastructure. Falling camera prices and better
	communication technology have pushed the number of installed cameras worldwide past
	one billion \cite{comparitech_2025}, and the way these systems are used has changed as
	well. Classical surveillance was largely passive: footage was stored and inspected only
	after an incident occurred \cite{sreenu_durai}. Advances in computer vision and deep
	learning now make it possible to process video streams automatically, shifting the
	emphasis from recording toward the rapid identification of relevant events.
	
	This shift responds to a concrete problem. The volume of video data far exceeds what
	human operators can analyze effectively. In classical monitoring centers a single
	operator watches dozens of streams at once, yet studies show that continuous monitoring
	is demanding and that attention and event-detection ability drop significantly after a
	relatively short period \cite{sulman_2008,nasholm_2014}. Because most footage shows
	ordinary, incident-free scenes, sustaining attention is even harder. Automatic video
	analysis, which can flag only the situations that require attention, is therefore a
	necessary complement to human operators.
	
	\subsection{Motivation}
	
	The gap between the amount of video generated and the limited attention of operators
	justifies intelligent surveillance in general, but it does not by itself justify a new
	system, since many commercial products already exist. Our motivation instead starts
	from the limitations of those products. Most commercially available solutions are
	\emph{closed} and \emph{proprietary}. This model creates strong vendor lock-in and high
	acquisition and licensing costs, but its most serious drawback is that these systems
	behave as a ``black box'': the algorithms make decisions in the background, and the user
	can neither inspect how they work nor tune their parameters \cite{burrell_2016}. Such
	rigidity makes the systems hard to adapt to specific situations or to integrate with
	existing equipment.
	
	In contrast, this work aims to show that a competitive intelligent surveillance system
	can be built on \emph{accessible, open, and flexible} principles. Accessibility comes
	from using only established open-source technologies, which removes licensing costs
	\cite{hoffmann_2024}; openness gives full transparency over the data-processing flow;
	and flexibility is achieved through a modular architecture in which detection algorithms
	run independently and can be enabled, disabled, or replaced with ease. The practical
	value of such an architecture spans several domains: in military settings an independent
	solution avoids sending sensitive data to external cloud servers \cite{shafee_2025};
	public institutions benefit from the absence of licensing costs and from auditability;
	and private users gain the freedom to customize functionality.
	
	\subsection{Problem Statement}
	
	The central problem addressed can be summarized as follows: \emph{how can one build an
		integrated surveillance system that analyzes images in real time and performs several
		distinct processing tasks simultaneously, while keeping response time low and offering an
		interface that is easy for human operators to use?} This question combines several
	distinct challenges. The first is the \emph{diversity and complexity of the analysis
		tasks}: an effective system must extract many types of information (identity analysis,
	demographic characterization, vehicle identification, behavior analysis, and threat
	detection) and, crucially, must do so \emph{simultaneously} on the same streams without
	the component modules competing excessively for hardware resources. The second is
	\emph{real-time operation}: to play a proactive role, a security system must analyze
	streams at a rate close to real capture, which imposes strict per-frame time budgets even
	though modern models are computationally heavy. The problem thus becomes one of
	\emph{architecture}: how to coordinate several complex models so that response time stays
	short on ordinary, affordable hardware. The third is the \emph{usefulness of the operator
		interface}: results must be presented so as to guide attention quickly toward real
	problems, with intelligent alarm handling that suppresses redundant notifications.
	
	\subsection{Contributions}
	
	The main contributions of this paper are:
	
	\begin{itemize}
		\item \textbf{A dedicated weapon detector.} A single-class YOLOv8m detector is fine-tuned
		on a merged, relabeled dataset built from two heterogeneous public sources, with non-weapon
		images strategically retained as negative examples to suppress false positives. The model
		reaches a mAP@0.5 of 0.947.
		
		\item \textbf{A custom action recognizer.} A SlowFast-R50 network is fine-tuned in two
		stages for the classes \{normal, fight, vandalism\}, reaching 94.33\% validation accuracy
		and providing a detector for a category that public solutions rarely address.
		
		\item \textbf{A new vandalism dataset.} A purpose-built video corpus of 614 manually
		validated vandalism clips is created for a category that public datasets under-represent,
		enabling the training of the action recognizer above.
		
		\item \textbf{A temporal event-validation architecture.} Rather than exposing raw
		per-frame detections, every detector is wrapped in a
		validation layer---multi-frame confirmation, confidence-weighted temporal voting, and
		cascaded filtering---that converts noisy detections into reliable events. Its benefit is
		quantified experimentally: for license plates, image enhancement plus IoU tracking and
		confidence-weighted voting raise exact-match accuracy from 66.7\% to 81.8\% on video; for fire
		and smoke, a five-level filtering cascade with temporal confirmation cuts the false-alarm rate
		from 52\% to 4\%; and for faces, an open-set decision rule with a double threshold and
		per-person spatial authorization yields 97\% identification accuracy with a 0\%
		false-acceptance rate.
		
		\item \textbf{A unified real-time multi-task pipeline} that runs the five heterogeneous tasks
		concurrently on the same streams. A seven-thread design with a ``last-available-frame'' fusion
		policy loads each model once, shares a single GPU, and keeps per-frame latency below 100~ms on
		commodity hardware, so display rate is bounded by the camera rather than by the slowest
		detector.
		
		\item \textbf{A systematic evaluation} of every module on independent public datasets (LFW~\cite{lfw2007},
		D-Fire~\cite{dfire2022}, FIRESENSE~\cite{firesense2017}, UCF-Crime~\cite{ucfcrime2018}), together with integrated end-to-end functional and stress
		testing that demonstrates correct real-time operation under maximum concurrent load.
	\end{itemize}
	
	The remainder of the paper is organized as follows. Section~\ref{sec:related} reviews
	commercial products, open-source and academic solutions, and the state of research in the
	relevant vision domains. Section~\ref{sec:methods} describes the multi-task
	architecture and the temporal event-validation layer that gives the system its operational
	reliability, with emphasis on the two custom-trained models. Section~\ref{sec:results} reports the experimental evaluation, and
	Section~\ref{sec:conclusion} concludes.

	In the interest of reproducibility, the complete source code supporting the findings of this paper are made publicly available at \url{https://github.com/DumitruEstera/ai-surveillance-system}.

	\section{Related Work}
	\label{sec:related}
	
	The development of surveillance has passed through several stages, each addressing the
	limitations of the previous generation: the transition from analog to IP cameras
	\cite{ip_surveillance}, the integration of Video Content Analysis (VCA)
	\cite{vca_optimization,vca_response}, and, most recently, the deep-learning revolution
	\cite{sreenu_durai}. Whereas classical VCA relied on manually programmed rules that were
	inflexible in complex scenes, convolutional neural networks learn directly from large data
	sets, classifying objects with high precision in real time. This section positions our
	system relative to commercial products, open-source and academic solutions, and the research
	literature.
	
	\subsection{Commercial Solutions}
	
	The market is dominated by robust commercial platforms that offer advanced capabilities but
	operate as closed systems with recurring costs. \emph{Verkada} \cite{verkada_overview} is a cloud-native platform recognized for ease of installation and centralized management; it performs edge analytics such as person and vehicle detection, facial-attribute and clothing-color recognition, license plate recognition (LPR), and unusual-behavior detection. Its main
	constraint is total dependence on the vendor's own cameras (vendor lock-in) and a total cost of
	ownership that grows through per-camera recurring licenses \cite{verkada_security}.
	\emph{Avigilon} \cite{avigilon_appearance} is a traditional video management system (VMS) leader whose flagship Appearance Search lets operators query people and vehicles across many streams;
	extracting maximum AI performance, however, typically requires Avigilon cameras or dedicated AI
	appliances, and budgets are inaccessible to small organizations. \emph{BriefCam} \cite{briefcam_datasheet} is a pure software solution famous for Video Synopsis and for advanced analytics (demographics, LPR, face recognition, behavioral analysis), but its complex models impose a
	major hardware barrier, requiring powerful on-premise servers with multiple enterprise GPUs
	\cite{briefcam_hardware}. Together these products confirm the high level that applied computer
	vision has reached, while also confirming the limitations---hardware dependence, high licensing
	costs, and lack of transparency---that motivate an open alternative.
	
	\subsection{Open-Source and Academic Solutions}
	
	The alternative to commercial systems is the open-source community and academic research. These
	initiatives remove licensing costs and hardware lock-in but often lack optimization for very
	specific use cases or a modern, end-user-ready software architecture. \emph{Frigate NVR} \cite{frigate_nvr} is a popular AI-native network video recorder that runs fully local inference and supports hardware accelerators; its default models, trained on general datasets such as COCO~\cite{coco2014},
	target general-purpose object categories, so specialized functionality requires integrating
	additional custom models, and it is oriented mainly toward smart-home monitoring.
	\emph{ZoneMinder} \cite{zoneminder_docs} is a mature VMS supporting many IP cameras via RTSP, but
	it uses a monolithic C++/PHP architecture in which AI-based recognition is bolted on through
	external scripts and webhooks, frequently triggered only after classical motion detection, which
	adds latency. Finally, a large body of academic work applies YOLO detectors to security tasks and
	reports excellent accuracy and frame rates for firearm detection from CCTV
	\cite{fierro2026,mandal2026}; yet most of this work stops at a proof-of-concept stage, offering
	isolated Python scripts that analyze pre-recorded videos or single frames without a backend that
	manages multiple RTSP connections, a database for alerting, or an interactive interface. Even more
	advanced IoT prototypes \cite{khan2026} remain dedicated to a narrow scenario rather than a generic
	end-to-end product. Our system moves beyond this stage by turning validated models into a complete
	end-to-end application built on a modern FastAPI/React stack.
	
	\subsection{State of Research in the Relevant Domains}
	
	Whereas the previous subsections analyzed integrated \emph{products}, this one takes a
	\emph{task}-oriented view of the four vision domains on which our system rests.
	
	\emph{Face recognition} has been redefined by angular-margin loss functions that directly
	optimize the separability of feature embeddings. The reference point is \emph{ArcFace}
	\cite{arcface2019}, whose additive angular margin produces highly discriminative embeddings;
	recent surveys \cite{facerec_review2025} confirm that convolutional networks trained with
	ArcFace-style losses remain dominant in practice because of their balance of accuracy, robustness
	to pose and lighting, and low inference cost, and comparative studies under degraded conditions
	\cite{arcface_facenet2026} match exactly the conditions of surveillance. Our face module (using
	the InsightFace \texttt{buffalo\_s} package, whose 512-dimensional embeddings are ArcFace-trained
	on Glint360K~\cite{partialfc2021}) adopts this established line but moves the contribution from the model to the system
	that exploits it: FAISS indexing~\cite{faiss2021}, a double-threshold decision rule, and per-person spatial
	authorization.
	
	\emph{Object detection} in real time is dominated by the YOLO family, which treats detection as a
	single regression problem and achieves an optimal speed--accuracy trade-off; recent reviews
	\cite{yolo_review2023,yolo_benchmark2024} document the transition to anchor-free designs (from
	YOLOv8 \cite{yolov8}) and NMS-free inference (YOLOv10~\cite{yolov10}), and note that medium-size variants offer the best
	capacity/cost ratio for real-time deployment. We use YOLO for three tasks: plate detection
	(YOLOv8), fire/smoke detection (YOLOv10), and weapon detection, for which---absent a sufficiently
	robust public model---we fine-tune a dedicated single-class YOLOv8m detector reaching mAP@0.5 of
	0.947, comparable to values reported in the CCTV firearm-detection literature
	\cite{fierro2026,mandal2026}. The key difference is that those works stop at validating an
	isolated model, whereas our detector operates as one module inside an integrated system that
	overlays additional filtering and decision layers.
	
	\emph{Human action recognition} in security is usually treated as violence or anomalous-behavior
	detection. The reference spatio-temporal architecture is \emph{SlowFast} \cite{slowfast2019},
	which separates a Slow branch (spatial semantics) from a Fast branch (fine motion); \emph{RWF-2000}
	\cite{rwf2000} has become a benchmark of 2000 surveillance clips, and systematic reviews
	\cite{violence_review2024} confirm the migration from hand-crafted descriptors to deep
	spatio-temporal architectures. Our action module fine-tunes SlowFast-R50 (pretrained on
	Kinetics-400~\cite{kinetics2017}) for \{normal, fight, vandalism\}, reaching 94.33\% accuracy. It draws \texttt{normal}
	and \texttt{fight} from RWF-2000 for comparability, but because \texttt{vandalism} is
	under-represented in public resources it builds a dedicated corpus of 614 manually validated clips,
	extending applicability to a relevant but poorly covered incident category.
	
	\emph{Automatic license plate recognition} (ALPR) is addressed by an established two-stage
	architecture: an object detector localizes the plate, and an OCR engine reads the text
	\cite{alpr_yolov8_2024,alpr_review2025}. YOLO variants dominate the detection stage, and
	general-purpose engines such as EasyOCR~\cite{easyocr} (CRAFT~\cite{craft2019} for text detection, CRNN~\cite{crnn2017} for recognition) are common
	for reading \cite{alpr_yolov8_2024}; a recurring theme is the difficulty of recognition under low
	resolution, motion blur, and uneven lighting, which demands additional image preprocessing. Our
	plate module adopts the same two-stage design but places its contribution in the processing chain
	built around the models: explicit super-resolution by bicubic interpolation, bilateral filtering,
	adaptive histogram equalization (CLAHE), edge sharpening, a color fallback, and, above all, spatial
	tracking with confidence-weighted temporal voting validated by a regular expression for the Romanian
	plate format---temporal stabilization that is rarely present in works evaluating ALPR on static
	images but is essential for reliable operation on a continuous stream.
	
	A common feature of this literature is that each vision task is typically studied, optimized, and
	evaluated \emph{in isolation}. The reference models we adopt are validated choices; our specific
	contribution lies in two complementary directions: adapting and extending these models where the
	literature has gaps (a dedicated weapon detector and a purpose-built vandalism dataset), and, most
	importantly, integrating all these tasks---treated separately in the literature---into a single
	concurrent processing flow that runs in real time on accessible hardware.
	
	\section{Methods}
	\label{sec:methods}
	
	\subsection{Multi-Task Architecture and Design Rationale}
	
\begin{figure*}[t]
		\centering
		\begin{tikzpicture}[
			font=\small,
			>={Stealth[length=2.2mm]},
			node distance=3.8mm,
			box/.style={draw, rounded corners=3pt, thick, align=center},
			src/.style={box, minimum height=1.0cm, minimum width=1.95cm, fill=black!6},
			det/.style={box, minimum height=1.32cm, minimum width=3.0cm, fill=blue!7},
			cust/.style={box, line width=1pt, draw=orange!80!black, minimum height=1.32cm, minimum width=3.0cm, fill=orange!16},
			fuse/.style={box, minimum height=1.15cm, minimum width=2.25cm, fill=blue!16},
			outnode/.style={box, minimum height=0.95cm, minimum width=2.6cm, fill=green!12},
			arr/.style={->, thick, draw=black!62},
			hdr/.style={font=\footnotesize\bfseries, text=black!62, align=center},
			swatch/.style={draw, thick, minimum width=4mm, minimum height=3mm, inner sep=0pt},
			]
			\node[det] (face) {\textbf{Face \& demographics}\\[2pt]\scriptsize InsightFace + FAISS\\[1pt]\scriptsize double-threshold decision};
			\node[det, below=of face] (plate) {\textbf{License plate}\\[2pt]\scriptsize YOLOv8 + EasyOCR\\[1pt]\scriptsize confidence-weighted voting};
			\node[det, below=of plate] (fire) {\textbf{Fire \& smoke}\\[2pt]\scriptsize YOLOv10\\[1pt]\scriptsize five-level cascade};
			\node[cust, below=of fire] (act) {\textbf{Action recognition}\\[2pt]\scriptsize SlowFast-R50 \emph{(custom-trained)}\\[1pt]\scriptsize clip-level, multi-frame};
			\node[cust, below=of act] (weap) {\textbf{Weapon detection}\\[2pt]\scriptsize YOLOv8m \emph{(custom-trained)}\\[1pt]\scriptsize multi-frame confirmation};
			
			\node[src, left=1.25cm of fire] (cap) {\textbf{Frame capture}\\[2pt]\scriptsize per-detector clone};
			\node[src, above=6mm of cap] (cam) {\textbf{Video streams}\\[2pt]\scriptsize webcam / RTSP};
			\draw[arr] (cam.south) -- (cap.north);
			\foreach \n in {face,plate,fire,act,weap}{\draw[arr] (cap.east) -- (\n.west);}
			
			\node[box, line width=1pt, draw=green!50!black, fill=green!12, align=center, text width=2.3cm,
			minimum width=2.6cm, minimum height=8.1cm, right=1.4cm of fire] (val)
			{\textbf{Temporal Event Validation}\\[6pt]
				\scriptsize multi-frame\\[-1pt]\scriptsize confirmation\\[5pt]
				\scriptsize confidence-weighted\\[-1pt]\scriptsize voting\\[5pt]
				\scriptsize cascaded filtering\\[5pt]
				\scriptsize IoU tracking};
			\foreach \n in {face,plate,fire,act,weap}{\draw[arr] (\n.east) -- (\n.east -| val.west);}
			
			\node[fuse, right=1.3cm of val] (fuse) {\textbf{Fusion}\\[2pt]\scriptsize last-available-frame};
			\draw[arr] (val.east) -- (fuse.west);
			\node[outnode, above right=0.85cm and 1.15cm of fuse] (alarm) {\textbf{Alarms + logs}\\[2pt]\scriptsize dedup, PostgreSQL};
			\node[outnode, below right=0.85cm and 1.15cm of fuse] (ws) {\textbf{WebSocket}\\[2pt]\scriptsize annotated frames};
			\draw[arr] (fuse.east) -- (alarm.west);
			\draw[arr] (fuse.east) -- (ws.west);
			
			\begin{scope}[on background layer]
				\node[draw, dashed, rounded corners, draw=black!45, fit=(face)(weap), inner sep=9pt] (gpubox) {};
			\end{scope}
			
			\coordinate (hy) at ($(val.north)+(0,7mm)$);
			\node[hdr] at (cap.center      |- hy) {(1) Input};
			\node[hdr] at (gpubox.north    |- hy) {(2) Concurrent multi-task\\ inference --- shared GPU};
			\node[hdr] at (val.center      |- hy) {(3) Temporal\\ validation};
			\node[hdr] at (fuse.center     |- hy) {(4) Fusion\\ \& output};
			
			\node[swatch, fill=blue!7, below=1.05cm of gpubox.south west, anchor=north west] (lg1) {};
			\node[right=1.5mm of lg1, font=\scriptsize, anchor=west] (lg1t) {off-the-shelf pretrained model};
			\node[swatch, line width=1pt, draw=orange!80!black, fill=orange!16, right=6mm of lg1t, anchor=west] (lg2) {};
			\node[right=1.5mm of lg2, font=\scriptsize, anchor=west] {custom-trained for this work};
		\end{tikzpicture}
		\caption{Overall architecture of the proposed framework. Each frame from the video streams is
			dispatched, in parallel, to five detectors that run concurrently on a single shared GPU (two of them,
			highlighted in orange---action recognition and weapon detection---are models trained specifically for
			this work).
			Rather than emitting raw per-frame detections, every detector feeds a \emph{temporal event-validation}
			layer---multi-frame confirmation, confidence-weighted voting, and cascaded filtering---that converts
			noisy detections into reliable events; this layer is a key contribution and the main source of
			the false-alarm reduction reported in Section~\ref{sec:results}. Validated events converge in a fusion
			stage that applies a ``last-available-frame'' policy, generates deduplicated alarms and logs, and
			broadcasts the annotated frame over WebSocket, decoupling the display rate from the slowest detector.}
		\label{fig:pipeline}
	\end{figure*}
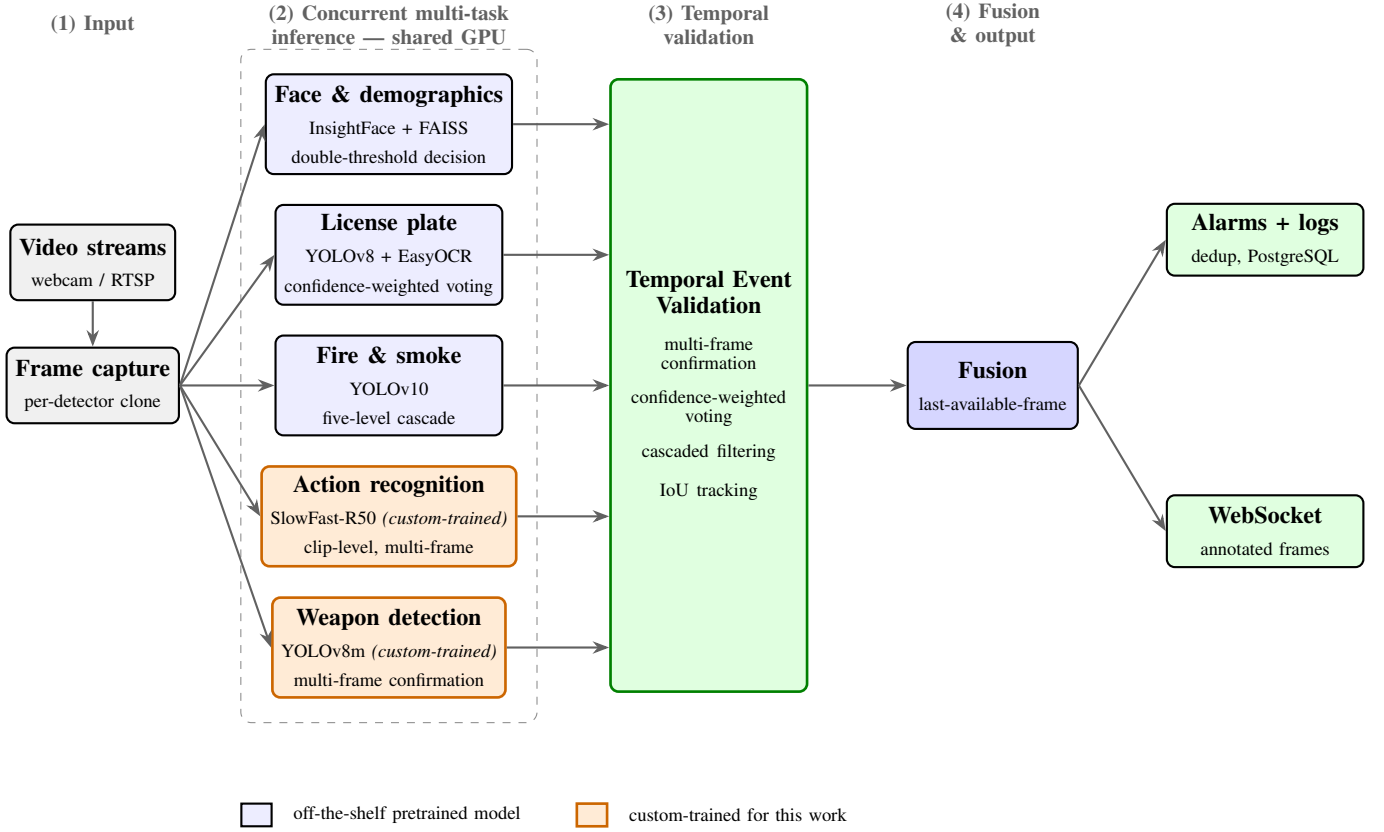

	The framework is built around a single principle: a security system is only as useful as it is
	\emph{reliable}, and the reliability of an AI detector on a continuous video stream is governed far
	more by how its per-frame outputs are validated over time than by its raw accuracy on isolated
	images. The architecture therefore has two coupled goals---running several heterogeneous vision
	tasks concurrently in real time, and validating each detector's output temporally before it becomes
	an event---and Fig.~\ref{fig:pipeline} shows how they fit together. Beyond real-time multi-task
	inference, the system also satisfies the usual operational requirements of a surveillance product
	(multi-camera streaming over RTSP/HTTP/MJPEG, runtime enabling/disabling of each detector, entity
	management, alarm deduplication, and exportable logs, exercised by an \emph{administrator} and an
	\emph{operator}), but these are supporting concerns rather than the focus of this paper.
	
	The first goal dictates a concurrent, in-process pipeline in which all detectors share a single GPU.
	This choice derives from three constraints: the AI models have inference times that span more than an
	order of magnitude (from a few milliseconds for YOLO plate detection up to tens of milliseconds for
	face recognition with demographics); each model's GPU initialization is expensive and must happen
	once; and the operator must be able to toggle detectors at runtime. The resulting pipeline uses seven
	daemon threads coupled only through bounded thread-safe queues.
	
	These threads---one capture thread, the five detector threads, and one fusion thread---isolate each
	responsibility and avoid fine-grained shared-variable synchronization; the capture thread clones each
	frame so every module works on its own copy. A key mechanism is the
	\emph{last-available-frame} policy in the fusion thread: waiting for all detectors would throttle
	throughput to the slowest one (action recognition delivers a result only about every two seconds),
	so the fusion thread instead consumes results non-blockingly, keeps the most recent result of each
	detector, and overlays them on the newest frame. Continuous-cadence detections (faces, plates,
	fire, weapons) are thus displayed at the camera rate, while the clip-cadence action label is carried
	forward until the next update. On saturation, input queues \emph{drop} incoming frames while output
	queues drop the \emph{oldest} result, ensuring that priority is always given to the most recent data.
	
	This parallel design was chosen over a sequential one for concrete, measurable reasons. In a sequential pipeline the latency perceived for \emph{any} detector is the sum of all active detectors' inference times; with the measured per-module latencies this sum is about 133~ms, dominated by the 83~ms of face recognition, which would drop the display rate below eight frames per second and exceed the 100~ms budget. In the parallel pipeline
	each module keeps its own latency and the last-available-frame mechanism makes display bounded by the camera rate. Runtime toggling of a detector reduces to simply not enqueuing frames for it---its thread stays blocked on \texttt{get()} at no CPU cost---and a recoverable failure in one detector does not propagate, since the fusion stage handles missing data transparently. A per-process microservice variant was rejected because it would multiply the costly GPU initialization (each process loading its own models, quickly exhausting VRAM) and add serialization latency, so a single
	process hosting all models on one GPU is preferable.
	
	Once a frame is ready for assembly, the annotations of the five modules are drawn in fixed layers over the image (face boxes and demographic labels first, then plates, confirmed fire, the recognized action label, and finally weapon boxes), producing a single annotated frame that concentrates all detections for that instant. In the same fusion stage, detections of interest are converted into two kinds of persistent records: exhaustive detection logs and proper alarms (critical events needing operator attention), the latter raised for unidentified persons, confirmed fire, non-normal
	actions, weapons, unauthorized vehicles, and unauthorized entry into a restricted zone. To avoid flooding the operator, a deduplication mechanism keys alarms by camera, event type, and subject, imposing a cooldown per unique key; including the subject in the key lets distinct alarms of the \emph{same} type coexist (two different unknown persons, a knife and a pistol, or two unauthorized vehicles) without being merged, and a redundant database check prevents duplicate cascades immediately after a server restart.
	
	Table~\ref{tab:models} summarizes, per task, the alternatives considered and the model chosen.
	Selection was guided by four criteria matched to a real-time surveillance system: concurrent
	real-time inference, robustness under real conditions (low resolution, uneven lighting, oblique
	angles, partial occlusion), availability of mature pretrained weights, and ease of integration into
	a concurrent Python architecture. Most modules reuse pretrained models in their original form; two
	were fine-tuned on purpose-built datasets (weapon detection and action recognition).
	
	\begin{table}[t]
		\caption{Model selection per computer vision task}
		\label{tab:models}
		\centering
		\renewcommand{\arraystretch}{1.25}
		\begin{tabularx}{\columnwidth}{@{}>{\raggedright\arraybackslash}p{1.9cm}>{\raggedright\arraybackslash}p{1.95cm}>{\raggedright\arraybackslash}X@{}}
			\toprule
			\textbf{Task} & \textbf{Chosen model} & \textbf{Main reason} \\
			\midrule
			Face detection \& recognition & InsightFace (\texttt{buffalo\_s}) & 512-d ArcFace embeddings; detection, recognition and demographics in one pass; efficient ONNX GPU inference. \\
			Age/gender & InsightFace \texttt{genderage} & Included in \texttt{buffalo\_s}; obtained in the same pass, no extra model. \\
			Emotion & Mini-Xception (FER)~\cite{minixception2017} & Compact CNN with negligible CPU latency, freeing the GPU. \\
			Object detection (plate / fire / weapon) & YOLOv8 / YOLOv10 & Single-stage real-time detectors; mature Ultralytics ecosystem; easy fine-tuning. \\
			Plate OCR & EasyOCR (CRAFT+CRNN) & Strong on low-resolution images; allowlist restricts the alphabet. \\
			Action recognition & SlowFast-R50 & Explicit Slow/Fast separation for motion; Kinetics-400 pretraining; real-time cost. \\
			Vector search & FAISS (\texttt{IndexFlatL2}) & Exact nearest-neighbor search; mature, native 512-d support. \\
			\bottomrule
		\end{tabularx}
	\end{table}
	
	\subsection{Temporal Event Validation}
	\label{sec:temporal}
	
	The central design choice of the framework is to never treat a raw per-frame detection as an event.
	A single frame is noise-prone: an OCR engine confuses visually similar characters, a fire detector
	fires on clouds, steam, or gray surfaces, and an object detector emits sporadic high-confidence false
	positives. Reported directly over continuous video, such per-frame outputs produce false-alarm rates
	that make an otherwise accurate detector operationally unusable. Each detector is therefore wrapped in a temporal event-validation layer that aggregates its outputs over several frames before any detection is allowed to raise an alarm. Four complementary mechanisms instantiate this principle:
	
	\begin{itemize}
		\item \emph{Multi-frame confirmation.} A candidate becomes a confirmed event only after it persists
		over several consecutive frames, associated across frames by IoU overlap. This is used by the
		fire/smoke, weapon, and action modules and is the dominant lever behind the fire false-alarm drop from
		52\% to 4\% reported in Section~\ref{sec:results}.
		\item \emph{Confidence-weighted temporal voting.} For license plates, per-frame readings are
		accumulated per track and the final text is decided by majority vote weighted by OCR confidence,
		correcting single-character errors that appear in only a minority of frames.
		\item \emph{Cascaded filtering.} For fire and smoke, a five-level cascade (per-class confidence, size
		plausibility, HSV color, temporal persistence, per-location cooldown) pre-filters candidates before
		temporal confirmation delivers the final decision.
		\item \emph{Open-set decision rules.} For faces, a double distance/confidence threshold rejects
		unknown identities outright, yielding a 0\% false-acceptance rate.
	\end{itemize}
	
	This layer also explains why static-image evaluation understates real performance: on an isolated
	frame these mechanisms operate without the temporal context they were designed for and appear
	conservative, whereas on video they sharply reduce false alarms at negligible cost in sensitivity
	(Section~\ref{sec:results}). The two detectors trained specifically for this work, described next, are
	consumed through exactly this layer.
	
	\subsection{Weapon Detection (Custom-Trained Model)}
	
	Unlike the reused-model modules, the weapon detector is trained entirely within this project, because
	no sufficiently robust public model exists for generic weapon detection in surveillance. Its data and
	training flow is structured in four scripts: download, relabel-and-merge, train, and evaluate.
	
	The model is trained to recognize a single generic class, \texttt{weapon}, covering firearms, edged
	weapons, and blunt objects. In a surveillance context what matters is the \emph{presence} of a weapon, not its exact type; grouping all weapons into one class reduces the task to binary detection, helps the model generalize, and removes confusions between subtypes that would trigger the same alert. The base model is YOLOv8m pretrained on COCO, chosen for its balance of accuracy and inference speed.
	
	Since no single public dataset offered sufficient coverage, two independent sources were combined:
	\emph{Dangerous Items}~\cite{dangerousitems} (Zenodo, 8478 images, five weapon classes) and \emph{SOHAS Weapon Detection}~\cite{sohas2020}
	(Roboflow, 5858 images, six classes of which two are weapons). A preprocessing step unifies them into a coherent YOLO structure. All weapon labels from both sources are remapped to class 0; the four non-weapon SOHAS classes (smartphone, banknote, wallet, card) are \emph{removed from the labels}, but their images are kept with empty label files. YOLOv8 treats such images as background: they contribute no positive targets, and any weapon detection on them is penalized as a false positive, so these images teach the model what is \emph{not} a weapon and reduce false positives on hand-held objects that resemble weapons. The train/validation/test split is inherited unchanged from the two pre-partitioned sources (no re-splitting is performed), and a source prefix (\texttt{zen\_}/\texttt{rf\_}) prevents filename collisions. The final combined dataset contains 14{,}336 images (roughly 70/15/15: 10{,}041 train, 2104 validation, 2191 test) and 12{,}592 weapon bounding boxes.
	
	The class distribution supports robust training. The Zenodo set is well balanced, with a similar number of annotations per class (from 1708 for \texttt{knife} to 1796 for \texttt{baseball\_bat}), preventing any single category from dominating; SOHAS adds a further 1510 \texttt{pistol} and 2277 \texttt{knife} annotations. Cumulated across both sources, edged weapons (\texttt{knife}, 3985 instances) and firearms (\texttt{pistol}, \texttt{rifle}, \texttt{gun}, 5018 instances) become the best-represented categories, matching the frequency of these threats in real scenarios. The four non-weapon SOHAS classes (smartphone, banknote, wallet, card; 2071 objects in total) are not
	discarded but reused as negative examples, teaching the model to avoid misclassifying hand-held objects with similar geometry.
	
	Training ran for 100 epochs (with early stopping, \texttt{patience}~=~20) at batch size 16 and $640\times640$ input, using the AdamW optimizer~\cite{adamw2019} with an initial learning rate of 0.001, a cosine schedule to a final factor of 0.01, weight decay 0.0005, and three warm-up epochs. Data augmentation combined HSV color jitter, geometric transforms (rotation, translation, scaling, horizontal flip), mosaic (disabled for the last ten epochs), and mixup. Training was performed on an NVIDIA A100 GPU. At inference, the module is consumed through the temporal event-validation layer of Section~\ref{sec:temporal}: it keeps only detections above a stricter secondary confidence filter (0.7) and requires multi-frame confirmation before raising an alert.
	
	\subsection{Human Action Recognition (Custom-Trained Model)}
	
	The most complex module classifies \emph{video sequences} (not single frames) into three behavioral
	categories: \texttt{normal}, \texttt{fight}, and \texttt{vandalism}. Recognizing actions requires modeling the temporal dimension, since the same image can belong to a hug or to an altercation depending on motion. The architecture is SlowFast-R50, whose parallel Slow (few frames, spatial semantics) and Fast (many frames, motion) pathways are joined by lateral connections, making it well suited to distinguishing actions by movement. The R50 backbone offers a good capacity/cost balance for real-time inference and is available pretrained on Kinetics-400.
	
	The fine-tuning dataset was built from three sources. The \texttt{normal} and \texttt{fight} classes are drawn from the established RWF-2000 surveillance dataset, ensuring comparability with published results. For \texttt{vandalism}, which public resources under-represent, a dedicated corpus was created: about 4800 candidate URLs were gathered via multilingual queries and web scraping from YouTube, followed by rigorous manual validation and segmentation, yielding 614 validated clips covering graffiti, window breaking, vehicle destruction, break-ins, and urban-furniture vandalism. In total the dataset comprises 2538 videos, split at the file level into 2097 for training and 441 for validation, with each video belonging exclusively to one partition.
	
	Preparation replaces the original Kinetics head \texttt{Linear(2304, 400)} with \texttt{Dropout(0.5)} followed by \texttt{Linear(2304, 3)}. Each clip is turned into a Slow/Fast tensor pair (8 and 32 frames respectively, cropped to $224\times224$, sampled from a 2.0~s window); the same input preparation is used identically at training and inference. Fine-tuning proceeds in two phases: for the first five epochs the backbone is frozen and only the new head is trained; from the sixth epoch the backbone is unfrozen and the whole model is trained with a reduced learning rate. At inference the running module accumulates frames in a ring buffer and runs one forward
	pass every 30 frames, applying softmax and reporting a non-normal action as an alert only above a confidence of 0.5, with a 5~s cooldown to avoid repeated notifications.
	
	\subsection{Face Recognition and Demographic Analysis}
	
	The face module is the core of access control. Unlike classical pipelines that chain a separate
	detector, embedding extractor, and demographic estimator, our implementation consolidates all these
	tasks into a single pass through the InsightFace \texttt{buffalo\_s} package, which bundles a face
	detector, a MobileFaceNet recognition network~\cite{mobilefacenet2018} producing 512-dimensional ArcFace embeddings, and a
	\texttt{genderage} head. For each frame, one call detects all faces and returns, per face, the
	bounding box, landmarks, the L2-normalized embedding, and age/gender estimates. To coexist with the
	concurrent PyTorch models~\cite{pytorch2019} on the same GPU, cuDNN convolution search is set to a heuristic mode that
	avoids the CUDA stream capture that would otherwise abort parallel PyTorch kernels; the emotion model
	(TensorFlow) is pinned to the CPU to prevent CUDA context conflicts.
	
	Each embedding is searched in a FAISS \texttt{IndexFlatL2} index, deterministically rebuilt from the
	database at startup. Identity is established by a \emph{double threshold}: a candidate is accepted as
	known only if the nearest neighbor has an $L_2$ distance below 1.0 \emph{and} a confidence
	(cosine similarity) above 0.4; otherwise the face is labeled \textit{Unknown}. Because embeddings are
	L2-normalized, $\text{cos\_sim} = 1 - L_2^2/2$. Beyond identification, the module verifies whether the
	person is authorized in the zone associated with the current camera and visually flags unauthorized
	individuals. A deliberate optimization conditions demographic analysis on the recognition result:
	age, gender, and emotion are computed and exposed \emph{only} for unknown faces, since a recognized
	employee's demographics are already known and re-computing them (an extra Mini-Xception CPU pass for
	emotion) would be redundant.
	
	The vector representations of known people are generated at enrollment: a person can register several face images to improve robustness, InsightFace runs on each, and the resulting embeddings are stored in the database and inserted into the FAISS index. Because the index is a volatile in-memory structure, it is rebuilt deterministically from the database at every startup, so losing the cache file never affects data integrity. Deleting a person triggers a rebuild of the
	index, and editing metadata (name, department, authorized zones) refreshes the authorization caches, keeping recognition and access decisions consistent at all times.
	
	\subsection{License Plate Recognition}
	
	The plate module follows the established two-stage design (YOLOv8 detection, EasyOCR reading) but
	adds a processing chain to cope with the difficult conditions typical of surveillance. Detected plate
	boxes are first expanded by 20\% in all directions to avoid clipping edge characters, then small
	crops are up-scaled toward a target width of 250~px using bicubic interpolation (scale capped at
	3.0 to limit noise amplification). The crop is then enhanced by a four-step routine tuned for EasyOCR:
	grayscale conversion; a bilateral filter that reduces noise while preserving character edges; CLAHE
	local histogram equalization to compensate for shadows and uneven lighting; and an unsharp mask to
	accentuate character edges after up-scaling. The image is deliberately \emph{not} binarized, since the
	CRNN engine performs better on continuous grayscale images. EasyOCR is called with an alphanumeric
	allowlist and parameters tuned for a single line of characters; if the enhanced reading is empty or
	its confidence falls below 0.4, a fallback re-runs OCR on the original color crop.
	
	Because single-frame OCR easily confuses visually similar characters (O/0, B/8) across frames, each
	detection is tracked over time. A \texttt{PlateTracker} associates new detections with existing tracks by IoU overlap, keeps a bounded history of recent readings per track, and ages out tracks that leave the field of view. Before entering the decision, each reading is validated against the Romanian plate format by a regular expression; only non-empty readings with confidence above 0.4 that match the format are stored. The final text for a track is obtained by \emph{confidence-weighted majority voting} over its valid history once at least three readings are collected. This temporal stabilization corrects single-character errors that appear in only a minority of frames.
	
	\subsection{Fire and Smoke Detection}
	
	The fire and smoke module wraps a pretrained YOLOv10 model (two classes, \texttt{fire} and \texttt{smoke}) in a five-level filtering cascade that turns a raw detector into an operational alarm source. Each candidate box produced by YOLOv10 must pass, in sequence: (1) a per-class confidence threshold (0.40 for fire, a stricter 0.55 for smoke, which is more prone to false positives from gray surfaces, clouds, or steam); (2) a size-plausibility check (between 0.3\% and 85\% of the frame area); (3) a heuristic color check in the HSV color space; (4) temporal confirmation through IoU tracking, so that a detection becomes \texttt{confirmed} only after persisting over several consecutive frames; and (5) a per-location alert cooldown that prevents repeated alerts for the same source. The cascade separates a \texttt{confirmed} state (a real event, drawn on the frame) from an \texttt{alert} state (the moment an alarm is raised), which sharply reduces false positives on continuous video, as shown in Section~\ref{sec:results}. This module is one of the three that rely on the multi-frame confirmation described in Section~\ref{sec:temporal}.
	
	\subsection{Implementation Note}
	
	The modules are hosted in a web application, which is not the focus of this paper: a monolithic FastAPI
	backend loads all models once, shares the GPU, exposes a REST API and a WebSocket channel for the annotated frames, and persists to PostgreSQL, from which the FAISS index is rebuilt at startup; a React front end with JWT authentication and role-based access control serves the operator interface.
	
	\section{Results}
	\label{sec:results}
	
	Each module was evaluated on public datasets and real surveillance footage, independent of the development environment. The evaluation targets not only raw accuracy but the properties that matter for a security application: the ability to reject unknown inputs, the false-alarm rate, and robustness under realistic conditions. Where a module includes its own processing flow---preprocessing, temporal voting, or multi-frame confirmation---testing explicitly compares the full system against the standalone model, to isolate the benefit of the additional validation logic. Unless otherwise stated, real-time measurements use an NVIDIA GeForce RTX~4060 Laptop GPU, an Intel Core Ultra~9 185H CPU, and 32~GB RAM.
	
	\subsection{Face Recognition}
	
	The face module was tested on LFW (Labeled Faces in the Wild) using an open-set 1:N identification protocol with impostor rejection, reproducing exactly the production decision rule (FAISS index, double threshold at $L_2<1.0$ and confidence $\geq0.4$). Known subjects are enrolled from one image subset and tested on a disjoint subset, while individuals appearing only once form the impostor set, whose correct response is always \textit{Unknown}. Images in which the detector does not find exactly one face are reported separately as detection failures and excluded from the accuracy computation, since the recognizer is not invoked.
	
	Two experiments were run. In a controlled experiment (12 known subjects, 20 impostors) the module reached 97.1\% identification accuracy with \emph{zero} identity confusions and 100\% impostor rejection (0\% false-acceptance rate, FAR). The three errors were conservative rejections of known subjects labeled \textit{Unknown}, which in practice only require a re-check rather than creating a vulnerability. A large-scale experiment extended the gallery to all 96 LFW subjects with at least 15 photos and included all 4069 single-appearance individuals as impostors (Table~\ref{tab:facial}). Accuracy held at 97.0\% (micro) and 96.9\% (macro), with 73 of the 96 subjects identified at
	100\%; again there were zero confusions and a 100\% rejection rate over thousands of impostors. The stability of the result under an eight-fold gallery expansion indicates good scalability. A threshold sweep confirmed the choice of $L_2=1.0$: rejection stays at 100\% across the range, while identification accuracy climbs from 83.0\% at a strict threshold of 0.8 and stabilizes near 97\% at 1.0, beyond which gains become marginal (98.8\% at 1.2).
	
	\begin{table}[t]
		\caption{Face recognition on LFW (decision threshold $L_2=1.0$)}
		\label{tab:facial}
		\centering
		\renewcommand{\arraystretch}{1.2}
		\begin{tabular}{@{}lcc@{}}
			\toprule
			\textbf{Metric} & \textbf{Controlled} & \textbf{Large-scale} \\
			\midrule
			Known subjects            & 12    & 96 \\
			Impostors                 & 20    & 4069 \\
			Test images (known)       & 120   & 1577 \\
			Identification acc.\ (micro) & 97.1\% & 97.0\% \\
			Identification acc.\ (macro) & ---    & 96.9\% \\
			Identity confusions       & 0     & 0 \\
			Impostor rejection        & 100\% & 100\% \\
			False acceptance (FAR)    & 0\%   & 0\% \\
			\bottomrule
		\end{tabular}
	\end{table}
	
	\subsection{License Plate Recognition}
	
	On 114 static test images (Table~\ref{tab:plate_static}), the full pipeline outperforms baseline EasyOCR on every metric, with the largest gain in the share of readings that conform to the valid Romanian format---decisive in operation, since the real system accepts a reading only if it matches the format. Reading the whole image without prior YOLO localization performs far worse (15.8\% exact match, CER 0.674), confirming that localization is the fundamental architectural decision of the module, as without it the OCR engine is distracted by irrelevant text. Restricting the analysis to the 101 images in which the plate was localized correctly makes the advantage even clearer: the pipeline reaches a 40.6\% exact-match rate, a CER of 0.244, and 59.4\% valid-format readings, against 34.7\%, 0.258, and 43.6\% for the raw crop reading.
	
	\begin{table}[t]
		\caption{License plate reading on 114 static test images}
		\label{tab:plate_static}
		\centering
		\renewcommand{\arraystretch}{1.2}
		\begin{tabularx}{\columnwidth}{@{}Xccc@{}}
			\toprule
			\textbf{Reading variant} & \textbf{Exact} & \textbf{Mean CER} & \textbf{Valid RO} \\
			\midrule
			Full pipeline            & 36.0\% & 0.330 & 52.6\% \\
			Raw EasyOCR on crop      & 30.7\% & 0.343 & 38.6\% \\
			Raw EasyOCR whole image  & 15.8\% & 0.674 & 21.1\% \\
			\bottomrule
		\end{tabularx}
	\end{table}
	
	The value of temporal voting is clearest on video. On 33 clips (Table~\ref{tab:plate_video}), temporal voting raises exact-match accuracy from 66.7\% (best single frame) to 81.8\%, recovering five plates misread on a single frame by correcting single-character errors through the majority vote of the frames in which the character was read correctly. The remaining failures correspond to plates misread in almost every frame (systematic font/condition confusions such as 5 vs.\ S), which no aggregation can fix. This result reflects the real operating regime and exceeds the pessimistic estimate from static images.
	
	\begin{table}[t]
		\caption{Effect of temporal voting on 33 video clips}
		\label{tab:plate_video}
		\centering
		\renewcommand{\arraystretch}{1.2}
		\begin{tabular}{@{}lcc@{}}
			\toprule
			\textbf{Method} & \textbf{Exact match} & \textbf{Mean CER} \\
			\midrule
			Single frame (no voting)      & 22/33 (66.7\%) & 0.173 \\
			Full system (temporal voting) & 27/33 (81.8\%) & 0.182 \\
			\bottomrule
		\end{tabular}
	\end{table}
	
	\subsection{Fire and Smoke Detection}
	
	Detection was evaluated at the binary alarm level: an image or clip is a positive prediction if the system reports at least one fire or smoke detection. On the D-Fire image set (4306 test images, 2005 negative; Table~\ref{tab:fire_img}), the per-frame filters cut the false-alarm rate from 19.4\% to 8.9\% but, as expected, also lower sensitivity (from 89.7\% to 53.6\%), because on an isolated frame they operate without the temporal validation for which they were designed---more than half (54\%) of the rejected positives are removed by the color filter, followed by the confidence (24\%) and size (22\%) filters. These filters are calibrated to pre-filter candidates, delegating the final decision to
	temporal validation on video.
	
	\begin{table}[t]
		\caption{Fire/smoke on D-Fire static images (alarm level)}
		\label{tab:fire_img}
		\centering
		\renewcommand{\arraystretch}{1.2}
		\begin{tabularx}{\columnwidth}{@{}Xccccc@{}}
			\toprule
			\textbf{Variant} & \textbf{Det.} & \textbf{FA} & \textbf{Prec.} & \textbf{F1} & \textbf{Acc.} \\
			\midrule
			Raw (YOLO only)          & 89.7\% & 19.4\% & 84.2\% & 0.87 & 85.5\% \\
			Pipeline (frame filters) & 53.6\% & 8.9\%  & 87.4\% & 0.66 & 71.1\% \\
			\bottomrule
		\end{tabularx}
	\end{table}
	
	That real operating regime is captured on the FIRESENSE video set (49 clips, 24 positive, 25 negative). Temporal confirmation keeps detection on positive clips near-perfect (23/24, 96\%) while cutting the false-alarm rate on negative clips from 52\% to just 4\% (Table~\ref{tab:fire_video}); by class, fire false alarms drop from 69\% to 6\% and smoke false alarms from 22\% to 0\%. A raw detector that alarms on more than half of harmless scenes would be practically unusable, so this reduction, at a negligible cost in sensitivity, is what turns the detector into a reliable operational module. This
	confirms that system effectiveness comes not only from the neural detector but equally from the integrated processing flow around it.
	
	\begin{table}[t]
		\caption{Fire/smoke on FIRESENSE video: effect of temporal confirmation}
		\label{tab:fire_video}
		\centering
		\renewcommand{\arraystretch}{1.2}
		\begin{tabularx}{\columnwidth}{@{}Xcc@{}}
			\toprule
			\textbf{Variant} & \textbf{Detection (pos.)} & \textbf{False alarm (neg.)} \\
			\midrule
			Raw (any frame)               & 24/24 (100\%) & 13/25 (52\%) \\
			Pipeline (temporal confirm.)  & 23/24 (96\%)  & 1/25 (4\%) \\
			\bottomrule
		\end{tabularx}
	\end{table}
	
	\subsection{Weapon Detection}
	
\begin{figure}[t]
		\centering
		\includegraphics[width=\columnwidth]{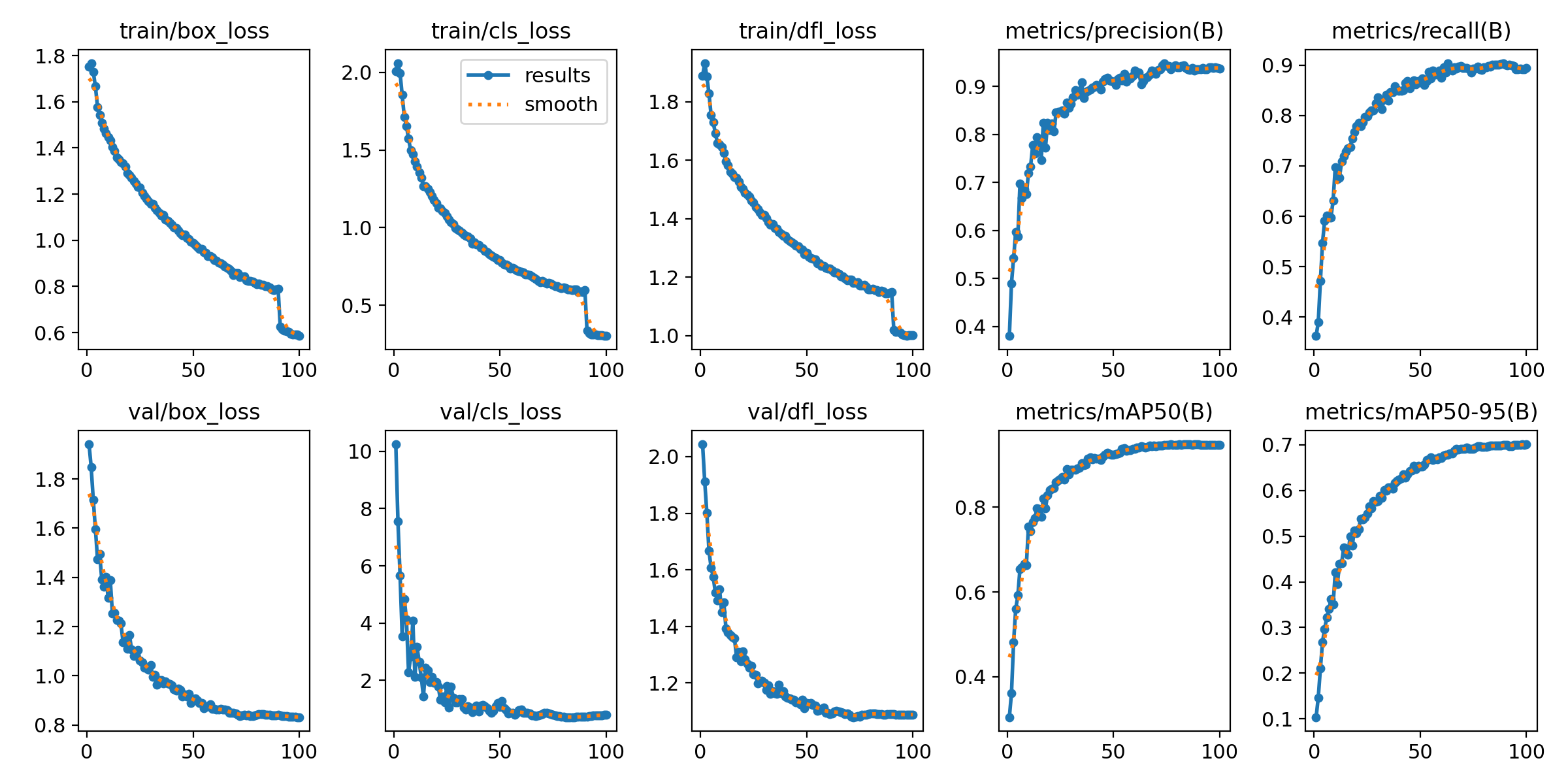}
		\caption{Training curves of the weapon detector over 100 epochs. Top row: training-set loss components
			(\texttt{box\_loss}, \texttt{cls\_loss}, \texttt{dfl\_loss}), precision and recall. Bottom row: the same validation-set loss components with mAP@0.5 and mAP@0.5:0.95. The abrupt training-loss drop near epoch 90 corresponds to disabling mosaic augmentation in the last ten epochs.}
		\label{fig:weapon_curves}
	\end{figure}

\begin{figure}[t]
		\centering
		\includegraphics[width=0.86\columnwidth]{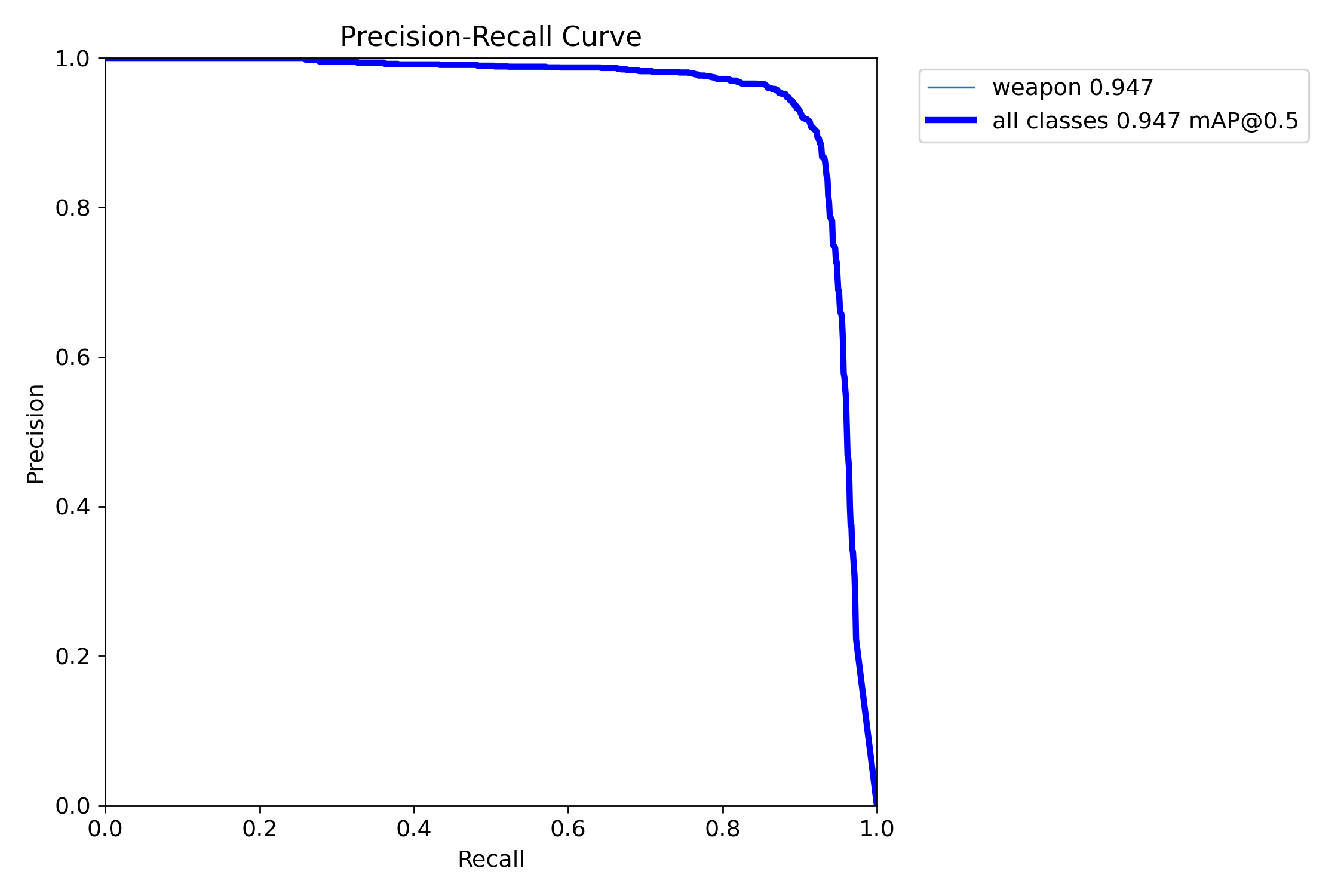}
		\caption{Precision--recall curve of the weapon detector on the validation set; the area under the curve equals mAP@0.5~=~0.947. Precision remains near its maximum up to a recall of about 0.9.}
		\label{fig:weapon_pr}
	\end{figure}

	The weapon detector was evaluated on a dedicated test partition kept completely separate from training and validation (2191 unseen images, 444 of them backgrounds without any weapon). It reaches a precision of 94.3\%, a recall of 90.2\%, and a mAP@0.5 of 0.943 (0.696 for mAP@0.5:0.95), practically identical to the validation figures (Table~\ref{tab:weapon}), confirming good generalization and the absence of overfitting. The training curves (Fig.~\ref{fig:weapon_curves}) show all loss components decreasing and stabilizing for both training and validation sets, with no systematic rise in validation loss; the abrupt drop near epoch 90 corresponds to disabling mosaic augmentation for the last ten epochs. Metrics rise fastest in the first 20--30 epochs, with precision climbing from about 0.38 to 0.94, recall from 0.36 to about 0.89, and mAP@0.5 from 0.30 to 0.947. The precision--recall curve (Fig.~\ref{fig:weapon_pr}) keeps precision close to 1.0 across most of the recall range, dropping only beyond a recall of about 0.9---a favorable trade-off for surveillance, where limiting false alarms is important. The gap between precision (94.3\%) and recall (90.2\%) shows a slightly conservative model that misses some weapons but produces few false detections; at inference this behavior is reinforced by a stricter secondary confidence filter and multi-frame confirmation, and the retained negative examples are reflected directly in the high precision.
	
	\begin{table}[t]
		\caption{Weapon detector metrics (validation and test)}
		\label{tab:weapon}
		\centering
		\renewcommand{\arraystretch}{1.2}
		\begin{tabular}{@{}lcc@{}}
			\toprule
			\textbf{Metric} & \textbf{Validation} & \textbf{Test} \\
			\midrule
			Precision        & 0.937 & 0.943 \\
			Recall           & 0.894 & 0.902 \\
			mAP@0.5          & 0.947 & 0.943 \\
			mAP@0.5:0.95     & 0.700 & 0.696 \\
			\bottomrule
		\end{tabular}
	\end{table}
	
	\subsection{Human Action Recognition}
	
\begin{figure}[t]
		\centering
		\includegraphics[width=0.72\columnwidth]{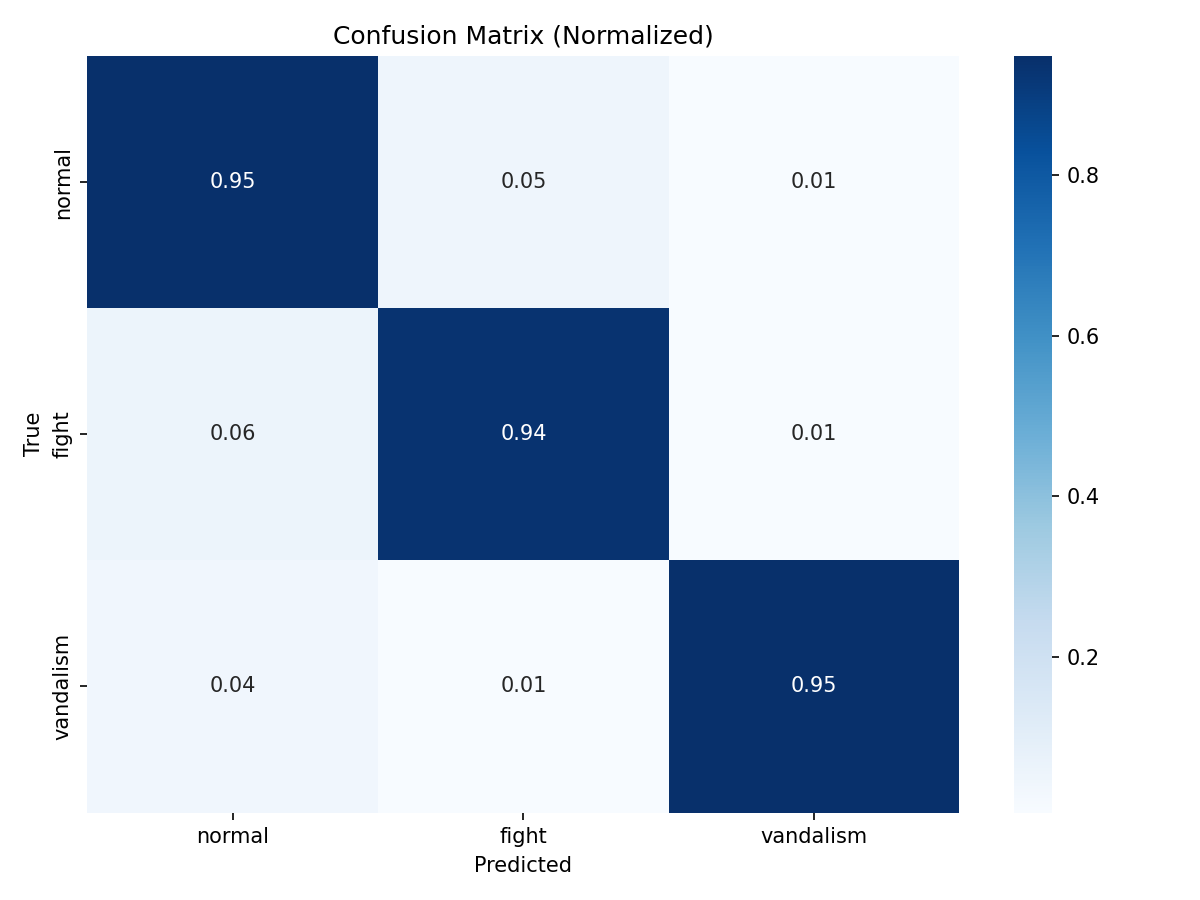}
		\caption{Normalized confusion matrix of the action recognizer on the validation partition. Rows are ground-truth classes, columns are predictions. The dominant diagonal confirms high accuracy; residual errors appear mostly as confusions between \texttt{normal} and \texttt{fight}.}
		\label{fig:har_cm}
	\end{figure}

\begin{figure}[t]
		\centering
		\includegraphics[width=\columnwidth]{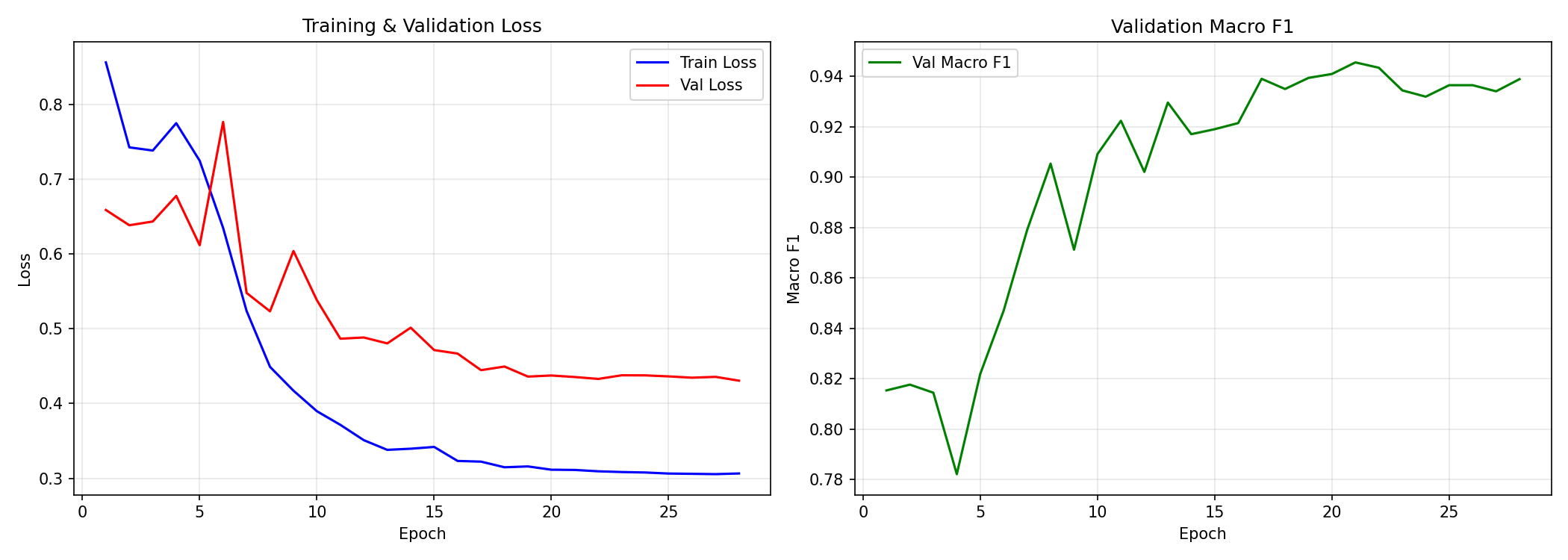}
		\caption{Training evolution of the action recognizer over 28 epochs. Left: training (blue) and validation (red) loss. Right: validation macro-F1. The perturbation around epochs 4--6 corresponds to unfreezing the backbone; the best weights were obtained at epoch 21, with early stopping at epoch 28.}
		\label{fig:har_curves}
	\end{figure}

	On the validation partition the fine-tuned SlowFast-R50 reaches 94.33\% accuracy and a macro-F1 of 0.9456 (macro precision 0.9478, macro recall 0.9436). Per-class results (Table~\ref{tab:har_val}) show the best performance on \texttt{vandalism} (precision 0.982, F1 0.965), the class for which the purpose-built dataset was created, indicating that the manually curated clips give a stable representation. The normalized confusion matrix (Fig.~\ref{fig:har_cm}) has a strong diagonal (0.95, 0.94, 0.95); residual errors occur mostly between \texttt{normal} and \texttt{fight}, both drawn from RWF-2000 and thus visually similar, while \texttt{vandalism} is confused with the other two classes in under 1\% of cases in each direction. The training curves (Fig.~\ref{fig:har_curves}) show training and validation loss decreasing steadily, with validation loss stabilizing around 0.43 and no overfitting; the transient rise around epochs 4--6 corresponds to unfreezing the backbone after
	the initial head-only phase, and the best weights were obtained at epoch 21, with early stopping at epoch 28.
	
	\begin{table}[t]
		\caption{Action recognition per-class metrics (validation)}
		\label{tab:har_val}
		\centering
		\renewcommand{\arraystretch}{1.2}
		\begin{tabular}{@{}lcccc@{}}
			\toprule
			\textbf{Class} & \textbf{Precision} & \textbf{Recall} & \textbf{F1} & \textbf{Support} \\
			\midrule
			\texttt{normal}    & 0.920 & 0.947 & 0.933 & 169 \\
			\texttt{fight}     & 0.942 & 0.935 & 0.939 & 155 \\
			\texttt{vandalism} & 0.982 & 0.949 & 0.965 & 117 \\
			\bottomrule
		\end{tabular}
	\end{table}
	
	To test generalization, the model was also evaluated on a subset of the real-surveillance UCF-Crime dataset (150 clips, 50 per class), using a sliding two-second window and an ``at least one'' aggregation that labels a clip as dangerous if any window is classified as fight or vandalism above threshold. The model reaches 86.7\% overall accuracy and a macro-F1 of 0.87 (Table~\ref{tab:har_ucf}). All 50 normal clips are classified correctly, generating no false alarms, while 82\% of fight clips and 78\% of vandalism clips correctly raise the corresponding alarm; the high precision (above 89\% for the dangerous classes) means that when the system flags an incident, the alert is almost always correct. Remaining errors are mainly missed detections when the action is filmed from a distance or is very short, plus occasional fight/vandalism confusions, which are understandable since both involve sudden aggressive movements.
	
	\begin{table}[t]
		\caption{Action recognition on UCF-Crime (150 clips, 50 per class)}
		\label{tab:har_ucf}
		\centering
		\renewcommand{\arraystretch}{1.2}
		\begin{tabular}{@{}lccc@{}}
			\toprule
			\textbf{Class} & \textbf{Recall} & \textbf{Precision} & \textbf{F1} \\
			\midrule
			Normal    & 100\% & 78\% & 0.88 \\
			Fight     & 82\%  & 89\% & 0.85 \\
			Vandalism & 78\%  & 98\% & 0.87 \\
			\midrule
			\multicolumn{4}{c}{Overall accuracy 86.7\% \quad$\vert$\quad macro-F1 0.87} \\
			\bottomrule
		\end{tabular}
	\end{table}
	
	\subsection{Integrated End-to-End Testing}
	
	Two complementary scenarios validated the integrated system. In a \emph{functional} scenario, a single clip in which each target appears in turn was processed with all detectors active (Table~\ref{tab:functional}). Every target was correctly identified: a known person (confidence 0.79) was distinguished from an unknown one (labeled \textit{Unknown} in 100\% of frames, with no false match), a plate was read exactly (CER~=~0), and the fire, weapon, and action events were all detected.
	Object-specific detectors (face, plate) reported near-zero latency, recognizing the target on the first processed frame, whereas fire (1.9~s) and weapon (3.2~s) latencies reflect the intentional multi-frame temporal confirmation, and the fight latency (about 2.9~s) reflects the periodic execution of the SlowFast module. Here latency is measured from the appearance of the object of interest in the clip to its first correct identification, relative to the moment the first frame enters
	processing.
	
	\begin{table}[t]
		\caption{Functional test of the integrated system (sequential scenario)}
		\label{tab:functional}
		\centering
		\renewcommand{\arraystretch}{1.2}
		\begin{tabularx}{\columnwidth}{@{}Xccc@{}}
			\toprule
			\textbf{Detector / target} & \textbf{OK} & \textbf{Latency} & \textbf{Conf.} \\
			\midrule
			Face --- known (identity)      & yes & 0.03~s & 0.79 \\
			Face --- unknown (\textit{Unknown}) & yes & 0.32~s & --- \\
			Plate --- \texttt{B112DGF} (CER 0) & yes & 0.06~s & 0.80 \\
			Action --- vandalism           & yes & 0.0~s  & 0.88 \\
			Action --- fight               & yes & 2.87~s & 0.92 \\
			Fire/smoke --- fire            & yes & 1.90~s & 0.81 \\
			Weapon --- weapon              & yes & 3.18~s & 0.89 \\
			\bottomrule
		\end{tabularx}
	\end{table}
	
	In a \emph{stress} scenario, a single six-quadrant split-screen clip forced all models to run simultaneously (Table~\ref{tab:stress}). The system processed about 10~frames per second drawn concurrently from all six streams; approximately 63\% of the native 30~fps stream is deliberately skipped to guarantee real-time processing---identical to a physical camera under overload---while losses at the internal queues stayed at only 3.6\%, confirming that the queues are adequately sized. The most expensive module is face recognition (63~ms per frame under load), followed by the YOLO
	detectors, with the SlowFast forward pass costing only 8~ms. A substantial resource reserve (about 62\% GPU capacity and 5.5~GB of free VRAM) remained, indicating headroom for further optimization or for processing more streams. For reference, Table~\ref{tab:latency} reports the per-module inference latencies measured in isolation on the same hardware.
	
	\begin{table}[t]
		\caption{Throughput and per-model latency under maximum concurrent load}
		\label{tab:stress}
		\centering
		\renewcommand{\arraystretch}{1.2}
		\begin{tabular}{@{}lr@{\hskip 1.2cm}lr@{}}
			\toprule
			\textbf{Throughput} & & \textbf{Latency (ms)} & \\
			\midrule
			Source FPS (native)      & 30.1 & Face recognition & 63 \\
			Effective processing FPS & $\approx$10 & Plate detection & 42 \\
			Frames arrived           & 2107 & Fire/smoke        & 42 \\
			Frames processed         & 747  & Weapon            & 39 \\
			Frames skipped (real time) & 63.2\% & Action (forward) & 8 \\
			Frames lost (full queues)  & 3.6\%  & & \\
			\bottomrule
		\end{tabular}
	\end{table}
	
	\begin{table}[t]
		\caption{Isolated per-module inference latency (RTX 4060 Laptop)}
		\label{tab:latency}
		\centering
		\renewcommand{\arraystretch}{1.2}
		\begin{tabularx}{\columnwidth}{@{}Xccc@{}}
			\toprule
			\textbf{Detector} & \textbf{Median (ms)} & \textbf{p95 (ms)} & \textbf{FPS} \\
			\midrule
			Plate (YOLOv8 + EasyOCR)\textsuperscript{a} & 4.0 & 4.7 & 249 \\
			Fire/smoke (YOLOv10)               & 24.6 & 25.6 & 41 \\
			Weapon (YOLOv8m)                   & 13.1 & 13.5 & 76 \\
			Action (SlowFast-R50, forward)     & 8.1 & 17.3 & 124 \\
			Face (InsightFace + FAISS + FER)   & 83.4 & 161.4 & 12 \\
			\bottomrule
			\multicolumn{4}{@{}l}{\footnotesize \textsuperscript{a}Median dominated by frames without a plate;} \\
			\multicolumn{4}{@{}l}{\footnotesize when EasyOCR fires, latency rises to $\sim$40--125~ms.}
		\end{tabularx}
	\end{table}
	
	\subsection{Discussion}
	
	A common thread runs through all these results: the processing steps that complement the AI models---temporal voting, multi-frame confirmation, and image enhancement---contribute decisively to real operating performance. Although these mechanisms can look overly conservative when evaluated on isolated images, on continuous video they sharply reduce false alarms without compromising sensitivity. Static-image performance can therefore be read as a conservative estimate of real
	behavior, while video results reflect deployment more faithfully. The integrated evaluation supports this conclusion, demonstrating real-time operation with a comfortable hardware reserve.
	
	Equally important is that these results were obtained on datasets and footage unseen during development, so the consistency between validation-time and independent-test performance (notably for the weapon detector and the action recognizer) indicates good generalization rather than dataset-specific tuning. The measured resource headroom under maximum concurrent load also confirms the central architectural claim: integrating five heterogeneous vision tasks in a single real-time flow on commodity hardware is feasible, and the bottleneck is the individual model latency rather than the fusion or coordination logic, which leaves clear room for scaling to additional streams.

	\section{Conclusion}
	\label{sec:conclusion}
	
	This paper presented a multi-task deep-learning framework for intelligent video surveillance that runs five computer vision tasks concurrently on the same streams---face recognition with demographic and emotion analysis, license plate recognition, fire and smoke detection, weapon detection, and human action recognition---and, above all, validates every detector's output temporally before it becomes an event. The tasks share a single GPU through a multi-threaded pipeline that sustains near-real-time processing, while the temporal event-validation layer is what turns individually noisy detectors into a reliable operational system.
	
	On the quantitative side, the two models trained within the project reached competitive performance: the SlowFast-R50 action recognizer obtained 94.33\% accuracy, and the YOLOv8m weapon detector obtained a mean average precision (mAP@0.5) of 0.947. For fire and smoke, the five-level filtering cascade and temporal validation greatly reduced the false-alarm rate relative to the raw detector output. The highest-value contributions are the temporal event-validation architecture that reduces false alarms across modules (the fire/smoke filtering cascade and multi-frame confirmation, plate confidence-weighted temporal voting, and the open-set face decision rule), the two models trained within the project, and the purpose-built vandalism dataset (614 video sequences collected, filtered, and edited manually) created for a category that public resources under-represent.
	
	The solution has limitations identified during evaluation. Architecturally, it runs on a single compute node with a single GPU and local persistence, which restricts scalability for very large numbers of simultaneous streams. At the module level, performance degrades in edge cases: heavily degraded or extremely angled plates, night lighting without infrared sensors, and very small or partially occluded faces; the action recognizer can also confuse similar classes such as fight and vandalism. The weapon detector shows a false-negative rate of about 7\%, acceptable in the general context but requiring caution in critical scenarios.
	
	Several directions remain for future work: cross-camera person re-identification for continuous subject tracking, fall detection and abandoned-object detection, migration to a distributed multi-GPU architecture, replacing Base64-over-WebSocket frame transport with WebRTC to reduce latency and bandwidth, integration with external alerting systems and a mobile app, and a search-by-image
	function over the archive. In conclusion, the work demonstrates the feasibility of integrating several complex computer vision tasks into a single, performant, and easy-to-use system, offering an open and extensible alternative to closed commercial solutions and a starting point for further development in intelligent video surveillance.
	

\end{document}